\documentclass[11pt,a4paper]{article}
\usepackage[hyperref]{emnlp-ijcnlp-2019}

\usepackage{times}
\usepackage{latexsym}
\usepackage{graphicx}
\usepackage{amsmath}
\usepackage{adjustbox}
\usepackage{amssymb}
\usepackage[normalem]{ulem}
\usepackage{stfloats}

\usepackage{array}
\newcolumntype{L}{>{\centering\arraybackslash}m{2cm}}

\usepackage{url}

\aclfinalcopy 

\newcommand\ftnote[1]{\footnote{\raggedright#1}}

\title{Enhancing AMR-to-Text Generation with Dual Graph Representations}

\author{Leonardo F. R. Ribeiro$^{\dag}$, Claire Gardent$^{\ddag}$ and Iryna Gurevych$^{\dag}$ \\\\
  $^{\dag}$Research Training Group AIPHES and UKP Lab, Technische Universit\"at Darmstadt\\
  \texttt{\href{https://www.ukp.tu-darmstadt.de}{www.ukp.tu-darmstadt.de}} \\
  \rule{0pt}{2.5ex}
  $^{\ddag}$CNRS/LORIA, Nancy, France \\
 \texttt{claire.gardent@loria.fr}
}

\date{}

\begin{document}
\maketitle
\begin{abstract}
Generating text from graph-based data, such as Abstract Meaning Representation (AMR), is a challenging task due to the inherent difficulty in how to properly encode the structure of a graph with labeled edges. To address this difficulty, we propose a novel \mbox{graph-to-sequence} model that encodes different but complementary perspectives of the structural information contained in the AMR graph. The model learns parallel top-down and bottom-up representations of nodes capturing contrasting views of the graph. We also investigate the use of different node message passing strategies, employing different state-of-the-art graph encoders to compute node representations based on incoming and outgoing perspectives. In our experiments, we demonstrate that the dual graph representation leads to improvements in \mbox{AMR-to-text} generation, achieving state-of-the-art results on two AMR datasets\ftnote{Code is available at \href{https://github.com/UKPLab/emnlp2019-dualgraph}{https://github.com/UKPLab/emnlp2019-dualgraph}}.
\end{abstract}


\section{Introduction}

Abstract Meaning Representation (AMR; \citet{banarescu-etal-2013-abstract}) is a linguistically-grounded semantic formalism that represents the meaning of a sentence as a rooted directed graph, where nodes are concepts and edges are semantic relations. As AMR abstracts away from surface word strings and syntactic structure producing a language neutral representation of meaning, its usage is beneficial in many semantic related NLP tasks, including text summarization \cite{liao-etal-2018-abstract} and machine translation \cite{song-etal-2019-semantic}.

The purpose of \mbox{AMR-to-text} generation is to produce a text which verbalises the meaning encoded by an input AMR graph. This is a challenging task as capturing the complex structural information stored in graph-based data is not trivial, as these are non-Euclidean structures, which implies that properties such as global parametrization, vector space structure, or shift-invariance do not hold \cite{geometric_deeplearning}. Recently, Graph Neural Networks (GNNs) have emerged as a powerful class of methods for learning effective graph latent representations \cite{xu2018how} and \mbox{graph-to-sequence} models have been applied to the task of \mbox{AMR-to-text} generation \cite{song-etal-acl2018,beck-etal-2018-acl2018,damonte_naacl18, dcgcnforgraph2seq19guo}.

In this paper, we propose a novel graph-to-sequence approach to AMR-to-text generation, which is inspired by pre-neural generation algorithms. These approaches explored alternative (top-down, bottom-up and mixed) traversals of the input graph and showed that a hybrid traversal combining both top-down (TD) and bottom-up (BU) information was best as this permits integrating both global constraints top-down from the input and local constraints bottom-up from the semantic heads \cite{shieber1990semantic,narayan-gardent-2012-structure}.

Similarly, we present an approach where the input graph is represented by two separate structures, each representing a different view of the graph. The nodes of these two structures are encoded using separate graph encoders so that each concept and relation in the input graph is assigned both a TD and a BU representation.

Our approach markedly differs from existing \mbox{graph-to-sequence} models for MR-to-Text generation \cite{marcheggiani-icnl18,beck-etal-2018-acl2018,damonte_naacl18} in that these approaches aggregate all the immediate neighborhood information of a node in a single representation. By exploiting parallel and complementary vector representations of the AMR graph, our approach eases the burden on the neural model in encoding nodes (concepts) and edges (relations) in a single vector representation. It also eliminates the need for additional positional information \cite{beck-etal-2018-acl2018} which is required when the same graph is used to encode both TD and BU information, thereby making the edges undirected.

Our main contributions are the following:
\begin{itemize}
    \item We present a novel architecture for \mbox{AMR-to-text} generation which explicitly encodes two separate TD and BU views of the input graph.
    \item We show that our approach outperforms recent \mbox{AMR-to-text} generation models on two datasets, including a model that leverages additional syntactic information \cite{cao_naacl19}.
    \item We compare the performance of three graph encoders, which have not been studied so far for \mbox{AMR-to-text} generation.
\end{itemize}


\section{Related Work}

Early works on \mbox{AMR-to-text} generation employ statistical methods \cite{flanigan-etal-2016-generation, pourdamghani-etal-2016-generating,castro-ferreira-etal-2017-linguistic} and apply linearization of the graph by means of a depth-first traversal.

Recent neural approaches have exhibited success by linearising the input graph and using a sequence-to-sequence architecture. \citet{konsas_17} achieve promising results on this
task. However, they strongly rely on named entities anonymisation. Anonymisation requires an ad hoc procedure for each new corpus. The matching procedure needs to match a rare input item correctly (e.g., ``United States of America'') with the corresponding part in the output text (e.g., ``USA'') which may be challenging and may result in incorrect or incomplete delexicalisations. In contrast, our approach omits anonymisation. Instead, we use a copy mechanism \cite{see-etal-2017-get}, a generic technique which is easy to integrate in the encoder-decoder framework and can be used independently of the particular domain and application. Our approach further differs from \citet{konsas_17} in that we build a dual TD/BU graph representation and use graph encoders to represent nodes.

\citet{cao_naacl19} factor the generation process leveraging syntactic information to improve the performance. However, they linearize both AMR and constituency graphs, which implies that important parts of the graphs cannot well be represented (e.g., coreference).

Several \mbox{graph-to-sequence} models have been proposed. \citet{marcheggiani-icnl18} show that explicitly encoding the structure of the graph is beneficial with respect to sequential encoding. They evaluate their model on two tasks, WebNLG \cite{gardent-etal-2017-webnlg} and SR11Deep \cite{belz-etal-2011-first}, but do not apply it to AMR benchmarks. \citet{song-etal-acl2018} and \citet{beck-etal-2018-acl2018} apply recurrent neural networks to directly encode AMR graphs. \citet{song-etal-acl2018} use a graph LSTM as the graph encoder, whereas \citet{beck-etal-2018-acl2018} develop a model based on GRUs. We go a step further in that direction by developing parallel encodings of graphs which are able to highlight different graph properties.

In a related task, \citet{rik_naacl19} propose an attention-based graph model that generates sentences from knowledge graphs. \citet{Schlichtkrull2018ModelingRD} use Graph Convolutional Networks (GCNs) to tackle the tasks of link prediction and entity classification on knowledge graphs.   

\citet{damonte_naacl18} show that off-the-shelf GCNs cannot achieve good performance for \mbox{AMR-to-text} generation. To tackle this issue, \citet{dcgcnforgraph2seq19guo} introduce dense connectivity to GNNs in order to integrate both local and global features, achieving good results on the task. Our work is related to \citet{damonte_naacl18}, that use stacking of GCN and LSTM layers to improve the model capacity and employ anonymization. However, our model is substantially different: (i) we learn dual representations capturing top-down and bottom-up adjuvant views of the graph, (ii) we employ more effective graph encoders (with different neighborhood aggregations) than GCNs and (iii) we employ copy and coverage mechanisms and do not resort to entity anonymization.

 \begin{figure*}
    \centering
    \includegraphics[width=1\textwidth]{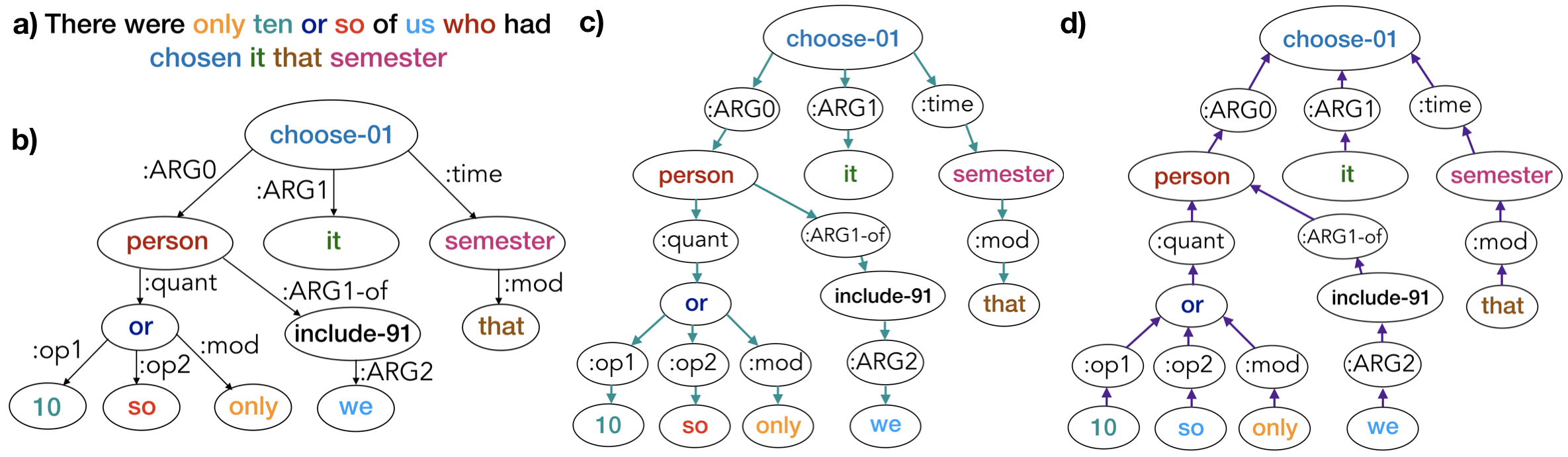}
    \caption{(a) an example sentence, (b) its original AMR graph ($G$) and different graph perspectives: (c) top-down ($G_t$) and (d) bottom-up ($G_b$).}
    \label{fig:rep_graphs}
\end{figure*}


\section{Graph-to-Sequence Model}
In this section, we describe (i) the representations of the graph adopted as inputs, (ii) the model architecture, including the Dual Graph Encoder and (iii) the GNNs employed as graph encoders.


\subsection{Graph Preparation} 
Let $G = (V, E, R)$ denote a rooted and directed AMR graph with nodes $v_i \in V$ and labeled edges $ (v_i, r, v_j) \in E$, where $r \in R$ is a relation type. Let $n = |V|$ and $m = |E|$ denote the numbers of nodes and edges, respectively. 

We convert each AMR graph into an unlabeled and connected bipartite graph $G_t = (V_t, E_t)$, transforming each labeled edge $(v_i, r, v_j) \in E$ into two unlabeled edges $(v_i, r) , (r, v_j) \in E_t$, with $|V_t| = n + m $ and $|E_t| = 2m $. This process, called Levi Transformation \cite{beck-etal-2018-acl2018}, turns original edges into nodes creating an unlabeled graph. For instance, the edge between {\fontfamily{qcr}\selectfont semester} and {\fontfamily{qcr}\selectfont that} with label {\fontfamily{qcr}\selectfont :mod} in Figure \ref{fig:rep_graphs}(b) is replaced by two edges and one node in \ref{fig:rep_graphs}(c): an edge between {\fontfamily{qcr}\selectfont semester}, and the new node {\fontfamily{qcr}\selectfont :mod} and another one between {\fontfamily{qcr}\selectfont :mod} and {\fontfamily{qcr}\selectfont that}. The new graph allows us to directly represent the relationships between nodes using embeddings. This enables us to encode label edge information using distinct message passing schemes employing different GNNs.

$G_t$ captures a TD view of the graph. We also create a BU view of the graph $G_b = (V_t, E_b)$, where each directed edge $e_k = (v_i, v_j) \in E_t$ becomes $e_k = (v_j, v_i) \in E_b$, that is, we reverse the direction of original edges. An example of a sentence, its AMR graph and the two new graphs $G_t$ and $G_b$ is shown in Figure~\ref{fig:rep_graphs}.

\subsection{Dual Graph Encoder}



We represent each node $v_i \in V_t$ with a node embedding $\mathbf{e}_i \in \mathbb{R}^d$, generated from the node label. In order to explicitly encode structural information, our encoder starts with two graph encoders, denoted by $GE_{t}$ and $GE_{b}$, that compute representations for nodes in $G_t$ and $G_b$, respectively.

Each $GE$ learns node representations based on the specific view of its particular graph, $G_t$ or $G_b$. Since $G_t$ and $G_b$ capture distinct perspectives of the graph structure, the information flow is propagated throughout TD and BU directions, respectively. In particular, for each node $v_i$, the $GE$ receives the node embeddings of $v_i$ and its neighbors, and computes its node representation:
 \begin{align*}
     \mathbf{h}^{t}_i =  GE_{t}(\{ \mathbf{e}_i, \mathbf{e}_j : j \in \mathcal{N}_t(i) \}), \\
     \mathbf{h}^{b}_i =  GE_{b}(\{ \mathbf{e}_i, \mathbf{e}_j : j \in \mathcal{N}_b(i) \}),
 \end{align*}
where $\mathcal{N}_t(i)$ and $\mathcal{N}_b(i)$ are the immediate incoming neighborhoods of $v_i$ in $G_t$ and $G_b$, respectively.

Each node $v_i$ is represented by two different hidden states, $\mathbf{h}^{t}_i$ and $\mathbf{h}^{b}_i$. Note that we learn two representations per relation and node of the original AMR graph. The hidden states $\mathbf{h}^{t}_i$ and $\mathbf{h}^{b}_i$, and embedding $\mathbf{e}_i$ contain different information regarding $v_i$. We concatenate them building a final node representation:
\begin{align*}
    \mathbf{r}_i = \big[ \, \mathbf{h}^{t}_i \, \Vert \, \mathbf{h}^{b}_i \, \Vert \, \mathbf{e}_i \, \big].
\end{align*}
This approach is similar to bidirectional RNNs \cite{Schuster97bidirectionalrecurrent}. Bidirectional RNNs benefit from left-to-right and right-to-left propagation. They learn the hidden representations separately and concatenate them at the end. We perform a similar encoding: first we learn TD and BU representations independently, and lastly, we concatenate them.

The final representation $\mathbf{r}_i$ is employed in a sequence input of a bidirectional LSTM. For each AMR graph, we generate a node sequence by depth-first traversal order. In particular, given a representation sequence from $\mathbf{r}_1$ to $\mathbf{r}_n$, the hidden forward and backward states of $\mathbf{r}_i$ are defined as:
\begin{align*}
\overrightarrow{\mathbf{h}}_i = \mathit{LSTM}_f(\mathbf{r}_i, \overrightarrow{\mathbf{h}}_{i-1}), \\
\overleftarrow{\mathbf{h}}_i = \mathit{LSTM}_b(\mathbf{r}_i, \overleftarrow{\mathbf{h}}_{i-1}),
\end{align*}
where $\mathit{LSTM}_f$ is a forward LSTM and $\mathit{LSTM}_b$ is a backward LSTM.  Note that, for the backward LSTM, we feed the reversed input as the order from $\mathbf{r}_n$ to $\mathbf{r}_1$. Lastly, we obtain the final hidden state by concatenating them as:
\begin{align*}
\mathbf{h}_i = [ \overrightarrow{\mathbf{h}}_i \, \Vert \, \overleftarrow{\mathbf{h}}_i ].
\end{align*}
The resulting hidden state $\mathbf{h}_i$ encodes the information of both preceding and following nodes.

\begin{figure}[t]
    \centering
    \includegraphics[width=.3\textwidth]{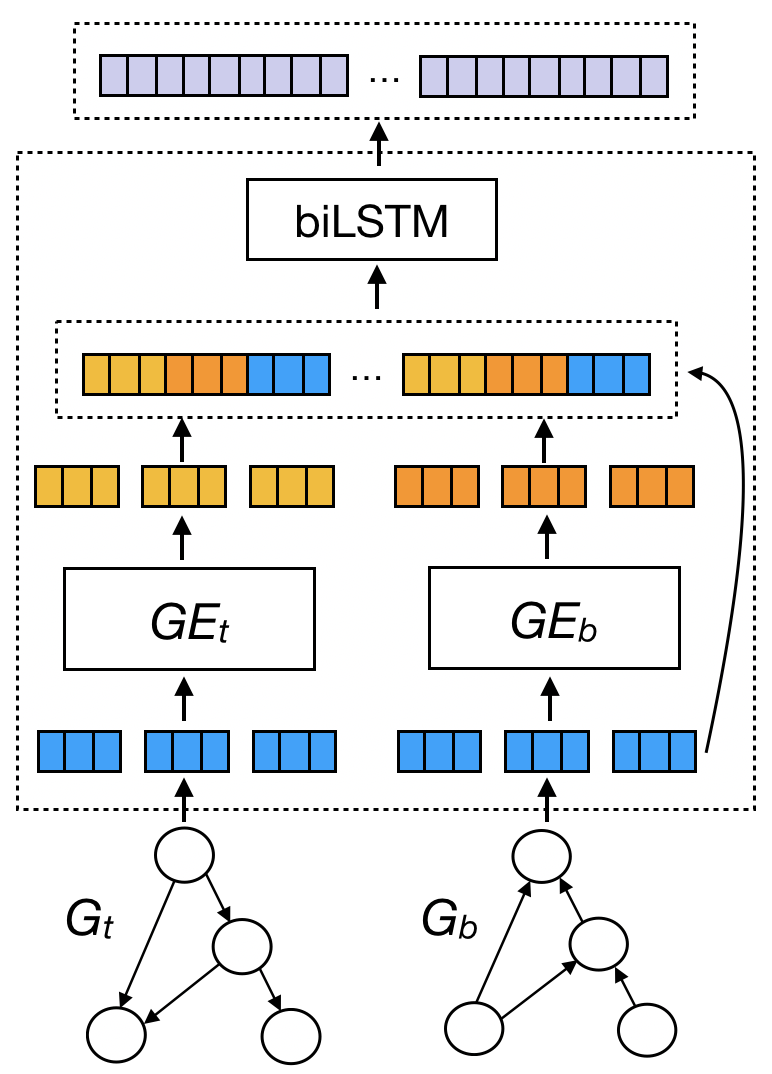}
    \caption{Dual Graph Encoder. The encoder receives the two graph views and generates structural node representations that are used by the decoder. Representations in blue, yellow and orange are $\mathbf{e}_i$, $\mathbf{h}^t_i$ and $\mathbf{h}^b_i$, respectively.}
    \label{fig:encoder}
\end{figure}

Stacking layers was demonstrated to be effective in \mbox{graph-to-sequence} approaches \cite{marcheggiani-icnl18, rik_naacl19, damonte_naacl18} and allows us to test for their contributions to the system performance more easily. We employ different GNNs for both graph encoders (Section \ref{sec:gnn}). Figure \ref{fig:encoder} shows the proposed encoder architecture. 


\subsection{Graph Neural Networks} \label{sec:gnn}

The $GE$s incorporate, in each node representation, structural information based on both views of the graph. We explore distinct strategies for neighborhood aggregation, adopting three GNNs: Gated Graph Neural Networks (GGNN, \citet{Li2016GatedGS}), Graph Attention Networks (GAT, \citet{velickovic2018graph}) and Graph Isomorphic Networks (GIN, \citet{xu2018how}). Each GNN employs a specific message passing scheme which allows capturing different nuances of structural information.

\paragraph{Gated Graph Neural Networks}
 GGNNs employ gated recurrent units to encode node representations, reducing the recurrence to a fixed number of steps. In particular, the $l$-th layer of a GGNN is calculated as:
\begin{align*}
        \mathbf{h}_i^{(l)} &= \mathit{GRU} \Big(
        \mathbf{h}_i^{(l-1)}, \: \sum_{j \in \mathcal{N}(i)}  \mathbf{W_1} \mathbf{h}_j^{(l-1)} \: \Big),
\end{align*}
where $\mathcal{N}(i)$ is the immediate neighborhood of $v_i$, $\mathbf{W_{1}}$ is a parameter and $\mathit{GRU}$ is a gated recurrent unit \cite{cho-etal-2014-learning}. Different from other GNNs, GGNNs use back-propagation through time (BPTT) to learn the parameters. GGNNs also do not require to constrain parameters to ensure convergence.

\paragraph{Graph Attention Networks}
GATs apply attentive mechanisms to improve the exploitation of non-trivial graph structure. They encode node representations by attending over their neighbors, following a self-attention strategy:
\begin{align*}
        \mathbf{h}_i^{(l)} &= \alpha_{i,i}\mathbf{W_2}\mathbf{h}_i^{(l-1)} +
        \sum_{j \in \mathcal{N}(i)} \alpha_{i,j}\mathbf{W_2}\mathbf{h}_j^{(l-1)},
\end{align*}
where attention coefficients $\alpha_{i,j}$ are computed as:
{\small
\begin{align*}
\alpha_{i,j} =
        \text{softmax} \left( \sigma\left(\mathbf{a}^{\top}
        [\mathbf{W_2}\mathbf{h}_i^{(l-1)} \, \Vert \, \mathbf{W_2}\mathbf{h}_j^{(l-1)}]
        \right)\right),
\end{align*}}
where $\sigma$ is the activation function and $\Vert$ denotes concatenation. $\mathbf{W_{2}}$ and $\mathbf{a}$ are model parameters. The virtue of the attention mechanism is its ability to focus on the most important parts of the node neighborhood. In order to learn attention weights in different perspectives, GATs can employ multi-head attentions.

\begin{table*}[t]
\begin{center}
 \begin{tabular}{p{6.5cm} c c c p{0.4cm} c c c}  
 \hline
 &  \multicolumn{3}{c}{\textbf{LDC2015E86}} & & \multicolumn{3}{c}{\textbf{LDC2017T10}} \\
 \hline
 training, dev and test instances & 16,833 & 1,368 & 1,371 & & 36,521 & 1,368 & 1,371 \\
 min, average and max graph diameter & 0 & 6.9 & 20 & & 0 & 6.7 & 20 \\
 min, average and max node degree & 0 & 2.1 & 18 & & 0 & 2.1 & 20 \\
 min, average and max number of nodes & 1 & 17.7 & 151 & & 1 & 16.8 & 151 \\
  min, average and max number of edges & 0 & 18.6 & 172 & & 0 & 17.7 & 172 \\
   number of DAG and non-DAG graphs & 18,679 & 893 & & & 37,284 & 1,976 &\\
 min, average and max length sentences & 1 & 21.3 & 225 & & 1 & 20.4 & 225 \\
 
  \hline
\end{tabular}
\caption{Data statistics of LDC2015E86 and LDC2017T10 datasets. The values are calculated for all splits (train, development and test sets). DAG stands for directed acyclic graph.}
\label{tab:datasets}
\end{center}
\end{table*}

\paragraph{Graph Isomorphic Networks}
GIN is a GNN as powerful as the Weisfeiler-Lehman (WL) graph isomorphism test \cite{weisfeiler} in representing isomorphic and non-isomorphic graphs with discrete attributes. Its $l$-th layer is defined as:
\begin{align*}
        \mathbf{h}_i^{(l)} &= h_{\mathbf{W}} \Big( \,
        \mathbf{h}_i^{(l-1)} + \sum_{j \in \mathcal{N}(i)} \mathbf{h}_j^{(l-1)} \, \Big),
\end{align*}
where $h_{\mathbf{W}}$ is a multi-layer perceptron (MLP). In contrast to other GNNs, which combine node feature with its aggregated neighborhood feature, GINs do not apply the combination step and simply aggregate the node along with its neighbors.

Each of these GNNs applies different approaches to learn structural features from graph data and has achieved impressive results on many graph-based tasks \cite{Li2016GatedGS, velickovic2018graph, xu2018how}.


\subsection{Decoder}

An attention-based unidirectional LSTM decoder is used to generate sentences, attending to the hidden representations of edges and nodes. In each step $t$, the decoder receives the word embedding of the previous word (during training, this is the previous word of the reference sentence; at test time it is the previously generated word), and has the decoder state $\mathbf{s}_t$. The attention distribution $\mathbf{a}^t$ is calculated as in \citet{see-etal-2017-get}:
\begin{align*}
        \mathbf{e}^t_{i} &= \mathbf{v} \cdot \text{tanh}(\mathbf{W}_h \mathbf{h}_i + \mathbf{W}_s \mathbf{s}_t + \mathbf{w}_c \mathbf{s}_c + \mathbf{b}), \\
        \mathbf{a}^t &= \text{softmax}(\mathbf{e}^t),
\end{align*}
where $\mathbf{s}_c$ is the coverage vector and $\mathbf{v}$, $\mathbf{W}_h$, $\mathbf{W}_s$, $\mathbf{w}_c$ and $\mathbf{b}$ are learnable parameters. The coverage vector is the accumulation of all attention distributions so far.

\paragraph{Copy and Coverage Mechanisms}
Previous works \cite{damonte_naacl18, cao_naacl19} use anonymization to handle names and rare words, alleviating the data sparsity. In contrast, we employ copy and coverage mechanisms to address out-of-vocabulary issues for rare target words and to avoid repetition \cite{see-etal-2017-get}.

The model is trained to optimize the negative log-likelihood:
\begin{align*}
        \mathcal{L} = - \sum_{t=1}^{|Y|} \log \, p(y_t | y_{1:t-1}, X; \theta),
\end{align*}
where $Y = y_1,\dots,y_{|Y|}$ is the sentence, $X$ is the AMR graph and $\theta$ represents the model parameters.





 

 \begin{figure}
    \centering
    \includegraphics[width=.5\textwidth]{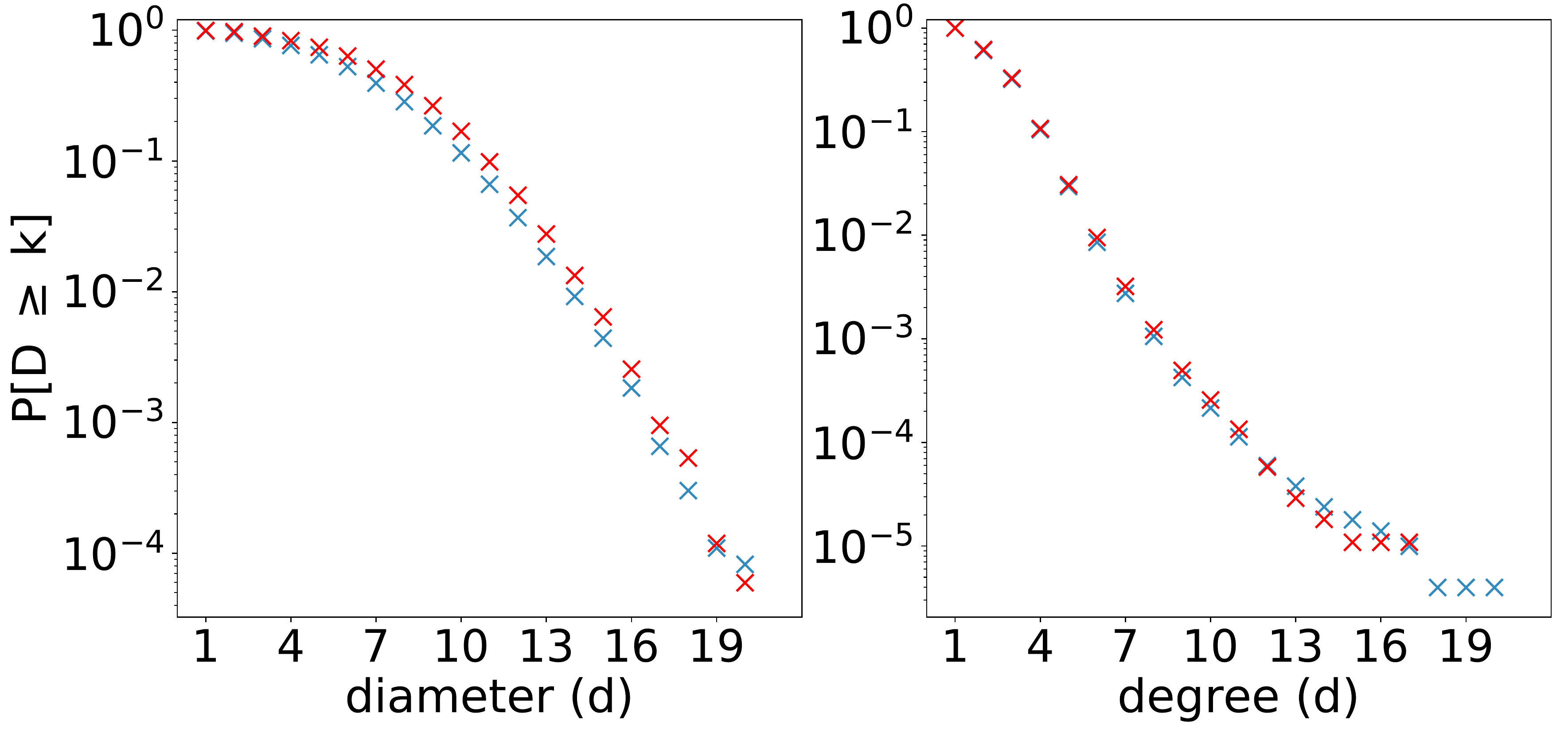}
    \caption{Distribution of the AMR graph diameter (left) and node degree (right) in the training set for LDC2015E86 (red) and LDC2017T10 (blue) datasets.}
    \label{fig:distdataset}
\end{figure}

\section{Data}

We use two AMR corpora, LDC2015E86 and LDC2017T10\ftnote{The datasets can be found at \href{https://amr.isi.edu/download.html}{https://amr.isi.edu/download.html}}. In these datasets, each instance contains an AMR graph and a sentence. Table~\ref{tab:datasets} shows the statistics for both datasets. Figure \ref{fig:distdataset} shows the distribution of the AMR graph diameters and node degrees for both datasets. The AMR graph structures are similar for most examples. Note that 90\% of AMR graphs in both datasets have the diameter less than or equal to 11 and 90\% of nodes have the degree of 4 or less. Very structurally similar graphs pose difficulty for the graph encoder by making it harder to learn the differences between their similar structures. Therefore, the word embeddings used as additional input play an important role in helping the model to deal with language information. That is one of the reasons why we concatenate this information in the node representation $\mathbf{r}_i$. 

\section{Experiments and Discussion}

 \begin{table}[t]
 \begin{tabular}{@{\hspace*{0.3mm}}p{3.37cm} @{\hspace*{0.3mm}}l @{\hspace*{3mm}}l@{\hspace*{0.3mm}}} 
 \hline
 \textbf{Model} & \textbf{BLEU} & \textbf{METEOR}   \\
 \hline
  \multicolumn{3}{c}{LDC2015E86} \\
 \hline
 {\normalsize Konstas et al. (2017)} & 22.00 & - \\ 
 {\normalsize Song et al. (2018)} & 23.28 & 30.10 \\ 
 Cao et al. (2019) & 23.50 & - \\
 {\normalsize Damonte et al.(2019)} & 24.40 & 23.60 \\
 Guo et al. (2019) & \textbf{25.70} & - \\
 \hline
  {\fontfamily{qcr}\selectfont S2S} & 22.55 {\small $\pm$ 0.17}  & 29.90 {\small $\pm$ 0.31} \\
  {\fontfamily{qcr}\selectfont G2S-GIN} &  22.93 {\small$ \pm$ 0.20} & 29.72 {\small $\pm$ 0.09} \\
  {\fontfamily{qcr}\selectfont G2S-GAT} &  23.42 {\small $\pm$ 0.16} & 29.87 {\small $\pm$ 0.14} \\
  {\fontfamily{qcr}\selectfont G2S-GGNN} & 24.32 {\small $\pm$ 0.16} & \textbf{30.53} {\small $\pm$ 0.30} \\
 \hline
 \multicolumn{3}{c}{LDC2017T10} \\
 \hline
 Back et al. (2018)  & 23.30 & - \\
 Song et al. (2018) & 24.86 & 31.56 \\ 
 {\normalsize Damonte et al.(2019)} & 24.54 & 24.07 \\
 Cao et al. (2019) & 26.80 & - \\
 Guo et al. (2019) & 27.60 & - \\
 \hline
  {\fontfamily{qcr}\selectfont S2S} & 22.73 {\small $\pm$ 0.18} & 30.15 {\small $\pm$ 0.14} \\
 {\fontfamily{qcr}\selectfont G2S-GIN} & 26.90 {\small $\pm$ 0.19} & 32.62 {\small $\pm$ 0.04} \\
 {\fontfamily{qcr}\selectfont G2S-GAT} & 26.72 {\small $\pm$ 0.20} & 32.52 {\small $\pm$ 0.02} \\
 {\fontfamily{qcr}\selectfont G2S-GGNN} & \textbf{27.87} {\small $\pm$ 0.15} & \textbf{33.21} {\small $\pm$ 0.15} \\
 \hline
\end{tabular}
\caption{BLEU and METEOR scores on the test set of LDC2015E86 and LDC2017T10 datasets.}
\label{tab:testresults}
\end{table}

\paragraph{Implementation Details}

We extract vocabularies (size of 20,000) from the training sets and initialize the node embeddings from GloVe word embeddings \cite{pennington-etal-2014-glove} on Common Crawl. Hyperparameters are tuned on the development set of the LDC2015E86 dataset. For GIN, GAT, and GGNN graph encoders, we set the number of layers to 2, 5 and 5, respectively. To regularize the model, during training we apply dropout \cite{Srivastava:2014:DSW:2627435.2670313} to the graph layers with a rate of 0.3. The graph encoder hidden vector sizes are set to 300 and hidden vector sizes for LSTMs are set to 900.

The models are trained for 30 epochs with early stopping based on the development BLEU score. For our models and the baseline, we used a two-layer LSTM decoder. We use Adam optimization \cite{kingma:adam} as the optimizer with an initial learning rate of 0.001 and 20 as the batch size. Beam search with the beam size of 5 is used for decoding.



 
 \paragraph{Results}
 We call the models {\fontfamily{qcr}\selectfont G2S-GIN} (isomorphic encoder), {\fontfamily{qcr}\selectfont G2S-GAT} (graph-attention encoder), and {\fontfamily{qcr}\selectfont G2S-GGNN} (gated-graph encoder), according to the graph encoder utilized. As a baseline ({\fontfamily{qcr}\selectfont S2S}), we train an attention-based encoder-decoder model with copy and coverage mechanisms, and use a linearized version of the graph generated by depth-first traversal order as input. We compare our models against several state-of-the-art results reported on the two datasets \cite{konsas_17,song-etal-acl2018, beck-etal-2018-acl2018, damonte_naacl18, cao_naacl19, dcgcnforgraph2seq19guo}.
 
  \begin{table}[t]
 \begin{tabular}{p{3.5cm} l p{1.5cm}} 
 \hline
 \textbf{Model} & \textbf{External} & \textbf{BLEU}   \\
 \hline
 {\normalsize Konstas et al. (2017)} & 200K & 27.40 \\
 {\normalsize Song et al. (2018)} & 200K & 28.20 \\
 {\normalsize Guo et al. (2019)} & 200K & 31.60 \\
 \hline
  {\fontfamily{qcr}\selectfont G2S-GGNN} & 200K & \textbf{32.23}  \\
 \hline
\end{tabular}
\caption{Results on LDC2015E86 test set when models are trained with additional Gigaword data.}
\label{tab:testresults-aug}
\end{table}
 
 We use both BLEU \cite{Papineni:2002:BMA:1073083.1073135} and METEOR \cite{Denkowski14meteoruniversal} as evaluation metrics\footnote{For BLEU, we use the multi-BLEU script from the MOSES decoder suite \cite{Koehn:2007:MOS:1557769.1557821}. For METEOR, we use the original \texttt{meteor-1.5.jar} script (\href{https://github.com/cmu-mtlab/meteor}{https://github.com/cmu-mtlab/meteor}).}. In order to mitigate the effects of random seeds, we report the averages for 4 training runs of each model along with their standard deviation. Table \ref{tab:testresults} shows the comparison between the proposed models, the baseline and other neural models on the test set of the two datasets. 

For both datasets, our approach substantially outperforms the baselines. In LDC2015E86, {\fontfamily{qcr}\selectfont G2S-GGNN} achieves a BLEU score of 24.32,  4.46\% higher than \citet{song-etal-acl2018}, who also use the copy mechanism. This indicates that our architecture can learn to generate better signals for text generation. On the same dataset, we have competitive results to \citet{damonte_naacl18}. However, we do not rely on preprocessing anonymisation not to lose semantic signals. In LDC2017T10, {\fontfamily{qcr}\selectfont G2S-GGNN} achieves a BLEU score of 27.87, which is 3.33 points higher than \citet{damonte_naacl18}, a state-of-the-art model that does not employ external information. We also have competitive results to \citet{dcgcnforgraph2seq19guo}, a very recent state-of-the-art model.

We also outperform \citet{cao_naacl19} improving BLEU scores by 3.48\% and 4.00\%, in LDC2015E86 and LDC2017T10, respectively. In contrast to their work, we do not rely on (i) leveraging supplementary syntactic information and (ii) we do not require an anonymization pre-processing step. {\fontfamily{qcr}\selectfont G2S-GIN} and {\fontfamily{qcr}\selectfont G2S-GAT} have comparable performance on both datasets. Interestingly, {\fontfamily{qcr}\selectfont G2S-GGNN} has better performance among our models. This suggests that graph encoders based on gating mechanisms are very effective in text generation models. We hypothesize that the gating mechanism can better capture long-distance dependencies between nodes far apart in the graph. 
%

  \begin{table}
 \begin{tabular}{p{3.4cm} @{\hspace*{2mm}}c @{\hspace*{2mm}}c @{\hspace*{2mm}}c@{\hspace*{2mm}}} 
 \hline
 \rule{0pt}{11pt} \textbf{Model} & {\small \textbf{BLEU}} & {\small \textbf{METEOR}} & \textbf{Size}  \\ 
 \hline
 {\fontfamily{qcr}\selectfont biLSTM} & 22.50 & 30.42  & 57.6M \\
 $GE_t$ + {\fontfamily{qcr}\selectfont biLSTM}  & 26.33 & 32.62 & 59.6M \\
 $GE_b$ + {\fontfamily{qcr}\selectfont biLSTM}  & 26.12 & 32.49 & 59.6M \\
 $GE_t$ + $GE_b$ + {\fontfamily{qcr}\selectfont\small biLSTM} & 27.37 & 33.30 & 61.7M  \\
 \hline
\end{tabular}
\caption{Results of the ablation study on the LDC2017T10 development set.}
\label{tab:ablation}
\end{table}

\paragraph{Additional Training Data} Following previous works \cite{konsas_17,song-etal-acl2018, dcgcnforgraph2seq19guo}, we also evaluate our models employing additional data from English Gigaword corpus \cite{Napoles:2012:AG:2391200.2391218}. We sample 200K Gigaword sentences and use JAMR\footnote{\href{https://github.com/jflanigan/jamr}{https://github.com/jflanigan/jamr}} \cite{flanigan-etal-2016-cmu} to parse them.
We follow the method of \citet{konsas_17}, which is fine-tuning the model on the LDC2015E86 training set after every epoch of pretraining on the Gigaword data. {\fontfamily{qcr}\selectfont G2S-GGNN} outperforms others with the same amount of Gigaword sentences (200K), achieving a 32.23 BLEU score, as shown in Table~\ref{tab:testresults-aug}. The results demonstrate that pretraining on automatically generated AMR graphs enhances the performance of our model.

\paragraph{Ablation Study}

In Table \ref{tab:ablation}, we report the results of an ablation study on the impact of each component of our model on the development set of LDC2017T10 dataset by removing the graph encoders. We also report the number of parameters (including embeddings) used in each model. The first thing we notice is the huge increase in metric scores (17\% in BLEU) when applying the graph encoder layer, as the neural model receives signals regarding the graph structure of the input. The dual representation helps the model with a different view of the graph, increasing BLEU and METEOR scores by 1.04 and 0.68 points, respectively. The complete model has slightly more parameters than the model without graph encoders (57.6M vs 61.7M). 

 \begin{table}
 \centering
 \begin{tabular}{@{\hspace*{0mm}}p{1.85cm} @{\hspace*{1.5mm}}l @{\hspace*{2mm}}l @{\hspace*{2mm}}l@{\hspace*{0mm}}} 
 \hline
 \rule{0pt}{12pt} \textbf{Model} & \multicolumn{3}{c}{\textbf{Graph Diameter}} \\
 \hline
 & 0-7 \,\,\,\,{\small \textbf{$\Delta$}} & 7-13 \,\,\,\,{\small \textbf{$\Delta$}} & 14-20 \,\,\,{\small \textbf{$\Delta$}} \\
 \hline
 {\fontfamily{qcr}\selectfont S2S} & 33.2 & 29.7 & 28.8 \\
 {\fontfamily{qcr}\selectfont G2S-GIN} & 35.2 {\small +6.0\%} & 31.8 {\small +7.4\%} & 31.5 {\small +9.2\%}  \\
 {\fontfamily{qcr}\selectfont G2S-GAT} & 35.1 {\small +5.9\%} & 32.0 {\small +7.8\%} & 31.5 {\small +9.51\%}  \\
 {\fontfamily{qcr}\selectfont G2S-GGNN} & 36.2 {\small +9.0\%} & 33.0 {\small +11.4\%} & 30.7 {\small +6.7\%}  \\
 
 \hline
\rule{0pt}{12pt} & \multicolumn{3}{c}{\textbf{Sentence Length}} \\ 
\hline
& {\small 0-20} \,\,\,\, {\small \textbf{$\Delta$}} & {\small 20-50} \,\, {\small \textbf{$\Delta$}} & {\small 50-240} \, {\small \textbf{$\Delta$}} \\
 \hline
 {\fontfamily{qcr}\selectfont S2S} & 34.9 & 29.9 & 25.1 \\
 {\fontfamily{qcr}\selectfont G2S-GIN} & 36.7 {\small +5.2\%} & 32.2 {\small +7.8\%} & 26.5 {\small +5.8\%}  \\
 {\fontfamily{qcr}\selectfont G2S-GAT} & 36.9 {\small +5.7\%} & 32.3 {\small +7.9\%} & 26.6 {\small +6.1\%}  \\
 {\fontfamily{qcr}\selectfont G2S-GGNN} & 37.9 {\small +8.5\%} & 33.3 {\small +11.2\%} & 26.9 {\small +6.8\%}  \\
  \hline
\rule{0pt}{12pt} & \multicolumn{3}{c}{\textbf{Max Node Out-degree}} \\
\hline
& {\small 0-3} \,\,\,\, {\small \textbf{$\Delta$}} & {\small 4-8} \,\, {\small \textbf{$\Delta$}} & {\small 9-18} \, {\small \textbf{$\Delta$}} \\
 \hline
 {\fontfamily{qcr}\selectfont S2S} & 31.7 & 30.0 & 23.9 \\
 {\fontfamily{qcr}\selectfont G2S-GIN} & 33.9 {\small +6.9\%} & 32.1 {\small +6.9\%} & 25.4 {\small +6.2\%}  \\
 {\fontfamily{qcr}\selectfont G2S-GAT} & 34.3 {\small +8.0\%} & 32.0 {\small +6.7\%} & 22.5 {\small -6.0\%}  \\
 {\fontfamily{qcr}\selectfont G2S-GGNN} & 35.0 {\small +10.3\%} & 33.1 {\small +10.4\%} & 22.2 {\small -7.3\%}  \\
  \hline
\end{tabular}
\caption{METEOR scores and differences to the {\fontfamily{qcr}\selectfont S2S}, in the LDC2017T10 test set, with respect to the graph diameter, sentence length and max node out-degree.}
\label{tab:stats-meteor}
\end{table}

\paragraph{Impact of Graph Size, Arity and Sentence Length}

The good overall performance on the datasets shows the superiority of using graph encoders and dual representations over the sequential encoder. However, we are also interested in estimating the performance of the models concerning different data properties. In order to evaluate how the models handle graph and sentence features, we perform an inspection based on different sizes of graph diameter, sentence length, and max node out-degree. Table~\ref{tab:stats-meteor} shows METEOR\footnote{METEOR score is used as it is a sentence-level metric.} scores for the LDC2017T10 dataset.

The performances of all models decrease as the diameters of the graphs increase. {\fontfamily{qcr}\selectfont G2S-GGNN} has a 17.9\% higher METEOR score in graphs with a diameter of at most 7 compared to graphs with diameters higher than 13.  This is expected as encoding a bigger graph (containing more information) is harder than encoding smaller graphs. Moreover, 71\% of the graphs in the training set have a diameter less than or equal to 7 and only 2\% have a diameter bigger than 13 (see Figure~\ref{fig:distdataset}). Since the models have fewer examples of bigger graphs to learn from, this also leads to worse performance when handling graphs with higher diameters.
We also investigate the performance with respect to the sentence length. The models have better results when handling sentences with 20 or fewer tokens. Longer sentences pose additional challenges to the models.

 \begin{table}[t]
 \centering
 \begin{tabular}{p{2cm} @{\hspace*{8mm}}c @{\hspace*{9mm}}c @{\hspace*{9mm}}c} 
 \hline
   \rule{0pt}{12pt} & \multicolumn{3}{c}{REF $\Rightarrow$ GEN} \\ 
  \hline
  \textbf{Model} & \textbf{ENT} & \textbf{CON} & \textbf{NEU} \\
 \hline
 {\fontfamily{qcr}\selectfont S2S} & 38.45 & 11.17 & 50.38 \\
 {\fontfamily{qcr}\selectfont G2S-GIN} & 49.78 & 9.80 & 40.42  \\
 {\fontfamily{qcr}\selectfont G2S-GAT} & 49.48 & 8.09 & 42.43  \\
 {\fontfamily{qcr}\selectfont G2S-GGNN} & 51.32 & 8.82 & 39.86  \\
 \hline
 \rule{0pt}{12pt} & \multicolumn{3}{c}{GEN $\Rightarrow$ REF} \\
 \hline
   \textbf{Model} & \textbf{ENT} & \textbf{CON} & \textbf{NEU} \\
 \hline
 {\fontfamily{qcr}\selectfont S2S} & 73.79 & 12.75 & 13.46 \\
 {\fontfamily{qcr}\selectfont G2S-GIN} & 76.27 & 10.65 & 13.08 \\
 {\fontfamily{qcr}\selectfont G2S-GAT} & 77.54  & 8.54 & 13.92  \\
 {\fontfamily{qcr}\selectfont G2S-GGNN} & 77.64  & 9.64  & 12.72  \\
\hline
\end{tabular}
\caption{Entailment (ENT), contradiction (CON) and neutral (NEU) average percentages for the LDC2017T10 test set. \textbf{(Top)} The premise and the hypothesis are the generated (GEN) and reference (REF) sentences, respectively. \textbf{(Bottom)} The hypothesis and the premise are the generated (GEN) and reference (REF) sentences, respectively.}
\label{tab:ent}
\end{table}

{\fontfamily{qcr}\selectfont G2S-GIN} has a better performance in handling graphs with node out-degrees higher than 9. This indicates that GINs can be employed in tasks where the distribution of node degrees has a long tail.
Surprisingly, {\fontfamily{qcr}\selectfont S2S} has a better performance than {\fontfamily{qcr}\selectfont G2S-GGNN} and {\fontfamily{qcr}\selectfont G2S-GAT} when handling graphs that contain high degree nodes.

\paragraph{Semantic Equivalence}
We perform an entailment experiment using BERT \cite{devlin2018bert} fine-tuned on the MultiNLI dataset \cite{williams-etal-2018-broad} as a NLI model. We are interested in exploring whether a generated sentence (hypothesis) is semantically \emph{entailed} by the reference sentence (premise). In a related text generation task, \citet{nli_summaries_acl} employ NLI models to rerank alternative predicted abstractive summaries. 

Nevertheless, uniquely verifying whether the reference (REF) entails the generated sentence (GEN) or vice-versa (GEN entails REF) is not sufficient. For example, suppose that \textit{``Today Jon walks''} is the REF and \textit{``Jon walks''} is the GEN. Even though REF entails GEN, GEN does not entail REF, that is, GEN is too general (missing information). Furthermore, suppose that \textit{``Jon walks''} is the REF and \textit{``Today Jon walks''} is the GEN, GEN entails REF but REF does not entail GEN, that is, GEN is too specific (added information). Therefore, in addition to verify whether the reference entails the generated sentence, we also verify whether the generated sentence entails the reference.

 \begin{figure}[t]
    \centering
    \includegraphics[width=.45\textwidth]{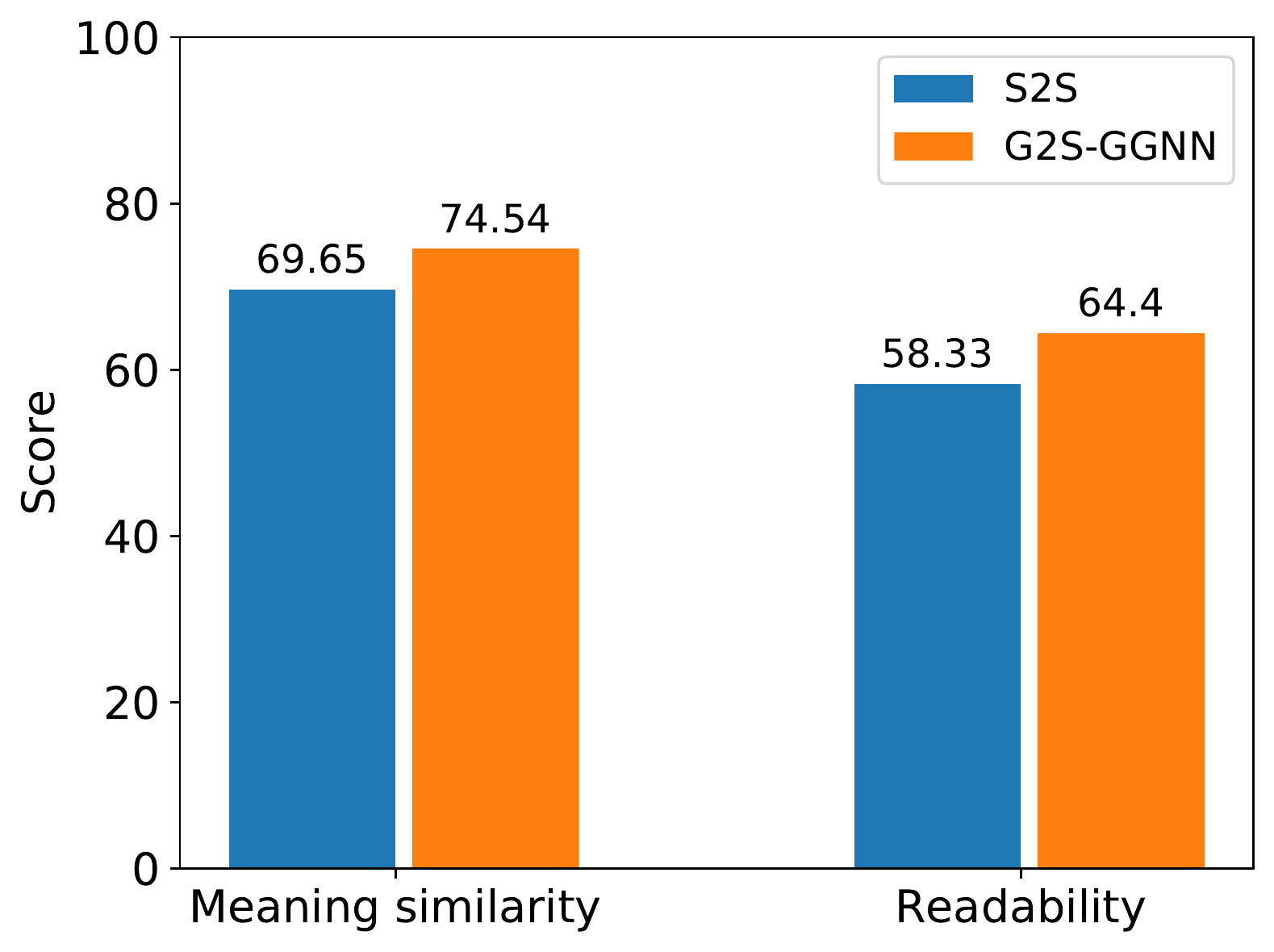}
    \caption{Human evaluation of the sentences generated by {\fontfamily{qcr}\selectfont S2S} and {\fontfamily{qcr}\selectfont G2S-GGNN} models. Results are statistically significant with $p < 0.05$, using Wilcoxon rank-sum test.}
    \label{fig:humaneval}
\end{figure}
Table~\ref{tab:ent} shows the average probabilities for entailment, contradiction and neutral classes on the LDC2017T10 test set. All {\fontfamily{qcr}\selectfont G2S} models have higher entailment compared to {\fontfamily{qcr}\selectfont S2S}. {\fontfamily{qcr}\selectfont G2S-GGNN} has 33.5\% and 5.2\% better entailment performances than {\fontfamily{qcr}\selectfont S2S}, when REF entails GEN and GEN entails REF, respectively. {\fontfamily{qcr}\selectfont G2S} models also generate sentences that contradict the reference sentences less. This suggests that our models are capable of capturing better semantic information from the graph generating outputs semantically related to the reference sentences.

\begin{table*}
\renewcommand{\arraystretch}{1.2}
\linespread{0.7}\selectfont\centering
 \begin{tabular}{p{2.75cm} p{12.5cm}} 
 \hline
\multicolumn{2}{p{15cm}}{{\footnotesize\fontfamily{qcr}\selectfont (a / agree :ARG0 (a2 / and :op1 (c / country :wiki China :name (n / name :op1 China)) :op2 (c2 / country :wiki Kyrgyzstan :name (n2 / name :op1 Kyrgyzstan))) :ARG1 (t / threaten-01 :ARG0 (a3 / and :op1 (t2 / terrorism) :op2 (s / separatism) :op3 (e / extremism)) :ARG2 (a4 / and :op1 (s3 / security :mod (r / region)) :op2 (s4 / stability :mod r)) :time (s2 / still) :ARG1-of (m / major-02)) :medium (c3 / communique :mod (j / joint))) }}\\
\hline
{\fontfamily{qcr}\selectfont GOLD} & {\small China and Kyrgyzstan agreed in a joint communique that terrorism, separatism and extremism still pose major threats to regional security and stability.}   \\
{\fontfamily{qcr}\selectfont S2S} & {\small In the joint communique, China and Kyrgyzstan still agreed to threaten terrorism, separatism, extremism and regional stability. }\\
Song et. al (2018) & {\small In a joint communique, China and Kyrgyzstan have agreed to still be a major threat to regional security, and regional stability.} \\
{\fontfamily{qcr}\selectfont G2S-GGNN} & {\small At a joint communique, China and Kyrgyzstan agreed that terrorism, separatism and extremism are still a major threat to region security and stability.} \\
 \hline
\end{tabular}
\caption{An example of an AMR graph and generated sentences. GOLD refers to the reference sentence.}
\label{tab:examples}
\end{table*}

 \begin{table}
 \centering
 \begin{tabular}{p{2cm} @{\hspace*{6mm}}c @{\hspace*{6mm}}c @{\hspace*{6mm}}c} 
 \hline
  \textbf{Model} & \textbf{ADDED} & \textbf{MISS} \\
 \hline
 {\fontfamily{qcr}\selectfont S2S} & 47.34 & 37.14 \\
 {\fontfamily{qcr}\selectfont G2S-GIN} & 48.67 & 33.64 \\
 {\fontfamily{qcr}\selectfont G2S-GAT} & 48.24 & 33.73 \\
 {\fontfamily{qcr}\selectfont G2S-GGNN} & 48.66 & 34.06 \\
 {\fontfamily{qcr}\selectfont GOLD} & 50.77 & 28.35 \\
 \hline

\end{tabular}
\caption{Fraction of elements in the output that are not present in the input (ADDED) and the fraction of elements in the input graph that are missing in the generated sentence (MISS), for the test set of LDC2017T10. The token lemmas are used in the comparison. {\fontfamily{qcr}\selectfont GOLD} refers to the reference sentences.}
\label{tab:stats-miss}
\end{table}

\paragraph{Human Evaluation}

To further assess the quality of the generated sentences, we conduct a human evaluation. We employ the \emph{Direct Assessment} (DA) method \cite{graham_baldwin_moffat_zobel_2017} via Amazon Mechanical Turk. Using the DA method inspired by \citet{mille-etal-2018-first}, we assess two quality criteria: (i) \emph{meaning similarity}: how close in meaning the generated text is to the gold sentence; and (ii) \emph{readability}:  how well the generated sentence reads (Is it good fluent English?).

We randomly select 100 sentences generated by {\fontfamily{qcr}\selectfont S2S} and {\fontfamily{qcr}\selectfont G2S-GGNN} and randomly assign them to HITs (following Mechanical Turk terminology).
Human workers rate the sentences according to meaning similarity and readability on a 0-100 rating scale. The tasks are executed separately and workers were first given brief instructions. For each sentence, we collect scores from 5 workers and average them. Models are ranked according to the mean of sentence-level scores. We apply a quality control step filtering workers who do not score some faked and known sentences properly.

Figure~\ref{fig:humaneval} shows the results. In both metrics, {\fontfamily{qcr}\selectfont G2S-GGNN} has better human scores for meaning similarity and readability, suggesting a higher quality of the generated sentences regarding {\fontfamily{qcr}\selectfont S2S}. The Pearson correlations between meaning similarity and readability scores, and METEOR\footnote{METEOR score is used as it is a sentence-level metric.} scores are 0.50 and 0.22, respectively.

\paragraph{Semantic Adequacy}
We also evaluate the semantic adequacy of our model (how well does the generated output match the input?) by comparing the number of added and missing tokens that occur in the generated versus reference sentences ({\fontfamily{qcr}\selectfont GOLD}). An added token is one that appears in the generated sentence but not in the input graph. Conversely, a missing token is one that occurs in the input but not in the output. In {\fontfamily{qcr}\selectfont GOLD}, added tokens are mostly function words while missing tokens are typically input concepts that differ from the output lemma. For instance, in Figure~\ref{fig:rep_graphs}, \textit{there} and \textit{of} are added tokens while \textit{person} is a missing token. 
As shown in Table~\ref{tab:stats-miss}, {\fontfamily{qcr}\selectfont G2S} approaches outperform the {\fontfamily{qcr}\selectfont S2S} baseline. {\fontfamily{qcr}\selectfont G2S-GIN} is closest to {\fontfamily{qcr}\selectfont GOLD} with respect to both metrics suggesting that this model is better able to generate novel words to construct the sentence and captures a larger range of concepts from the input AMR graph, covering more information.

\paragraph{Manual Inspection}
Table \ref{tab:examples} shows sentences generated by {\fontfamily{qcr}\selectfont S2S}, \citet{song-etal-acl2018}, {\fontfamily{qcr}\selectfont G2S-GAT}, and the reference sentence. The example shows that our approach correctly verbalises the subject of the embedded clause \textit{``China and ... agreed that \uline{terrorism, separatism and extremism}$_{ SUBJ}$ ... pose major threats to ...''}, while {\fontfamily{qcr}\selectfont S2S} and \citet{song-etal-acl2018} are fooled by the fact that \textit{agree} frequently takes  an infinitival argument which shares its subject (\textit{``\uline{China ...}$_{ SUBJ}$ agreed to threaten / have agreed to be a major threat''}).
While this is a single example, it suggests that dual encoding enhances the model ability to take into account the dependencies and the graph structure information, rather than the frequency of n-grams.



\section{Conclusion}
\label{sec:conclusion}

We have studied the problem of generating text from AMR graphs. We introduced a novel architecture that explicitly encodes two parallel and adjuvant representations of the graph (top-down and bottom-up). We showed that our approach outperforms state-of-the-art results in \mbox{AMR-to-text} generation. We provided an extensive evaluation of our models and demonstrated that they are able to achieve the best performance. In the future, we will consider integrating deep generative graph models to express probabilistic dependencies among AMR nodes and edges.



\section*{Acknowledgments}
This work has been supported by the German Research Foundation as part of the Research Training Group Adaptive Preparation of Information from Heterogeneous Sources (AIPHES) under grant No. GRK 1994/1.

\bibliography{emnlp-ijcnlp-2019}

\begin{thebibliography}{40}
\expandafter\ifx\csname natexlab\endcsname\relax\def\natexlab#1{#1}\fi

\bibitem[{Banarescu et~al.(2013)Banarescu, Bonial, Cai, Georgescu, Griffitt,
  Hermjakob, Knight, Koehn, Palmer, and
  Schneider}]{banarescu-etal-2013-abstract}
Laura Banarescu, Claire Bonial, Shu Cai, Madalina Georgescu, Kira Griffitt, Ulf
  Hermjakob, Kevin Knight, Philipp Koehn, Martha Palmer, and Nathan Schneider.
  2013.
\newblock \href {https://www.aclweb.org/anthology/W13-2322} {Abstract meaning
  representation for sembanking}.
\newblock In \emph{Proceedings of the 7th Linguistic Annotation Workshop and
  Interoperability with Discourse}, pages 178--186, Sofia, Bulgaria.
  Association for Computational Linguistics.

\bibitem[{Beck et~al.(2018)Beck, Haffari, and Cohn}]{beck-etal-2018-acl2018}
Daniel Beck, Gholamreza Haffari, and Trevor Cohn. 2018.
\newblock \href {https://www.aclweb.org/anthology/P18-1026} {Graph-to-sequence
  learning using gated graph neural networks}.
\newblock In \emph{Proceedings of the 56th Annual Meeting of the Association
  for Computational Linguistics (Volume 1: Long Papers)}, pages 273--283,
  Melbourne, Australia. Association for Computational Linguistics.

\bibitem[{Belz et~al.(2011)Belz, White, Espinosa, Kow, Hogan, and
  Stent}]{belz-etal-2011-first}
Anja Belz, Mike White, Dominic Espinosa, Eric Kow, Deirdre Hogan, and Amanda
  Stent. 2011.
\newblock \href {https://www.aclweb.org/anthology/W11-2832} {The first surface
  realisation shared task: Overview and evaluation results}.
\newblock In \emph{Proceedings of the Generation Challenges Session at the 13th
  {E}uropean Workshop on Natural Language Generation}, pages 217--226, Nancy,
  France. Association for Computational Linguistics.

\bibitem[{{Bronstein} et~al.(2017){Bronstein}, {Bruna}, {LeCun}, {Szlam}, and
  {Vandergheynst}}]{geometric_deeplearning}
Michael~M. {Bronstein}, Joan {Bruna}, Yann {LeCun}, Arthur {Szlam}, and Pierre
  {Vandergheynst}. 2017.
\newblock \href {https://doi.org/10.1109/MSP.2017.2693418} {Geometric deep
  learning: Going beyond euclidean data}.
\newblock \emph{IEEE Signal Processing Magazine}, 34(4):18--42.

\bibitem[{Cao and Clark(2019)}]{cao_naacl19}
Kris Cao and Stephen Clark. 2019.
\newblock \href {https://doi.org/10.18653/v1/N19-1223} {Factorising {AMR}
  generation through syntax}.
\newblock In \emph{Proceedings of the 2019 Conference of the North {A}merican
  Chapter of the Association for Computational Linguistics: Human Language
  Technologies, Volume 1 (Long and Short Papers)}, pages 2157--2163,
  Minneapolis, Minnesota. Association for Computational Linguistics.

\bibitem[{Castro~Ferreira et~al.(2017)Castro~Ferreira, Calixto, Wubben, and
  Krahmer}]{castro-ferreira-etal-2017-linguistic}
Thiago Castro~Ferreira, Iacer Calixto, Sander Wubben, and Emiel Krahmer. 2017.
\newblock \href {https://doi.org/10.18653/v1/W17-3501} {Linguistic realisation
  as machine translation: Comparing different {MT} models for {AMR}-to-text
  generation}.
\newblock In \emph{Proceedings of the 10th International Conference on Natural
  Language Generation}, pages 1--10, Santiago de Compostela, Spain. Association
  for Computational Linguistics.

\bibitem[{Cho et~al.(2014)Cho, van Merrienboer, Gulcehre, Bahdanau, Bougares,
  Schwenk, and Bengio}]{cho-etal-2014-learning}
Kyunghyun Cho, Bart van Merrienboer, Caglar Gulcehre, Dzmitry Bahdanau, Fethi
  Bougares, Holger Schwenk, and Yoshua Bengio. 2014.
\newblock \href {https://doi.org/10.3115/v1/D14-1179} {Learning phrase
  representations using {RNN} encoder{--}decoder for statistical machine
  translation}.
\newblock In \emph{Proceedings of the 2014 Conference on Empirical Methods in
  Natural Language Processing ({EMNLP})}, pages 1724--1734, Doha, Qatar.
  Association for Computational Linguistics.

\bibitem[{Damonte and Cohen(2019)}]{damonte_naacl18}
Marco Damonte and Shay~B. Cohen. 2019.
\newblock \href {https://doi.org/10.18653/v1/N19-1366} {Structural neural
  encoders for {AMR}-to-text generation}.
\newblock In \emph{Proceedings of the 2019 Conference of the North {A}merican
  Chapter of the Association for Computational Linguistics: Human Language
  Technologies, Volume 1 (Long and Short Papers)}, pages 3649--3658,
  Minneapolis, Minnesota. Association for Computational Linguistics.

\bibitem[{Denkowski and Lavie(2014)}]{Denkowski14meteoruniversal}
Michael Denkowski and Alon Lavie. 2014.
\newblock \href {https://doi.org/10.3115/v1/W14-3348} {Meteor universal:
  Language specific translation evaluation for any target language}.
\newblock In \emph{Proceedings of the Ninth Workshop on Statistical Machine
  Translation}, pages 376--380, Baltimore, Maryland, USA. Association for
  Computational Linguistics.

\bibitem[{Devlin et~al.(2019)Devlin, Chang, Lee, and
  Toutanova}]{devlin2018bert}
Jacob Devlin, Ming-Wei Chang, Kenton Lee, and Kristina Toutanova. 2019.
\newblock \href {https://doi.org/10.18653/v1/N19-1423} {{BERT}: Pre-training of
  deep bidirectional transformers for language understanding}.
\newblock In \emph{Proceedings of the 2019 Conference of the North {A}merican
  Chapter of the Association for Computational Linguistics: Human Language
  Technologies, Volume 1 (Long and Short Papers)}, pages 4171--4186,
  Minneapolis, Minnesota. Association for Computational Linguistics.

\bibitem[{Falke et~al.(2019)Falke, Ribeiro, Utama, Dagan, and
  Gurevych}]{nli_summaries_acl}
Tobias Falke, Leonardo F.~R. Ribeiro, Prasetya~Ajie Utama, Ido Dagan, and Iryna
  Gurevych. 2019.
\newblock \href {https://www.aclweb.org/anthology/P19-1213} {Ranking generated
  summaries by correctness: An interesting but challenging application for
  natural language inference}.
\newblock In \emph{Proceedings of the 57th Annual Meeting of the Association
  for Computational Linguistics}, pages 2214--2220, Florence, Italy.
  Association for Computational Linguistics.

\bibitem[{Flanigan et~al.(2016{\natexlab{a}})Flanigan, Dyer, Smith, and
  Carbonell}]{flanigan-etal-2016-cmu}
Jeffrey Flanigan, Chris Dyer, Noah~A. Smith, and Jaime Carbonell.
  2016{\natexlab{a}}.
\newblock \href {https://doi.org/10.18653/v1/S16-1186} {{CMU} at
  {S}em{E}val-2016 task 8: Graph-based {AMR} parsing with infinite ramp loss}.
\newblock In \emph{Proceedings of the 10th International Workshop on Semantic
  Evaluation ({S}em{E}val-2016)}, pages 1202--1206, San Diego, California.
  Association for Computational Linguistics.

\bibitem[{Flanigan et~al.(2016{\natexlab{b}})Flanigan, Dyer, Smith, and
  Carbonell}]{flanigan-etal-2016-generation}
Jeffrey Flanigan, Chris Dyer, Noah~A. Smith, and Jaime Carbonell.
  2016{\natexlab{b}}.
\newblock \href {https://doi.org/10.18653/v1/N16-1087} {Generation from
  abstract meaning representation using tree transducers}.
\newblock In \emph{Proceedings of the 2016 Conference of the North {A}merican
  Chapter of the Association for Computational Linguistics: Human Language
  Technologies}, pages 731--739, San Diego, California. Association for
  Computational Linguistics.

\bibitem[{Gardent et~al.(2017)Gardent, Shimorina, Narayan, and
  Perez-Beltrachini}]{gardent-etal-2017-webnlg}
Claire Gardent, Anastasia Shimorina, Shashi Narayan, and Laura
  Perez-Beltrachini. 2017.
\newblock \href {https://doi.org/10.18653/v1/W17-3518} {The {W}eb{NLG}
  challenge: Generating text from {RDF} data}.
\newblock In \emph{Proceedings of the 10th International Conference on Natural
  Language Generation}, pages 124--133, Santiago de Compostela, Spain.
  Association for Computational Linguistics.

\bibitem[{Graham et~al.(2017)Graham, Baldwin, Moffat, and
  Zobel}]{graham_baldwin_moffat_zobel_2017}
Yvette Graham, Timohy Baldwin, Alistair Moffat, and Justin Zobel. 2017.
\newblock \href {https://doi.org/10.1017/S1351324915000339} {Can machine
  translation systems be evaluated by the crowd alone}.
\newblock \emph{Natural Language Engineering}, 23(1):3–--30.

\bibitem[{Guo et~al.(2019)Guo, Zhang, Teng, and Lu}]{dcgcnforgraph2seq19guo}
Zhijiang Guo, Yan Zhang, Zhiyang Teng, and Wei Lu. 2019.
\newblock \href {https://doi.org/10.1162/tacl_a_00269} {Densely connected graph
  convolutional networks for graph-to-sequence learning}.
\newblock \emph{Transactions of the Association for Computational Linguistics},
  7:297--312.

\bibitem[{Kingma and Ba(2015)}]{kingma:adam}
Diederick~P Kingma and Jimmy Ba. 2015.
\newblock Adam: A method for stochastic optimization.
\newblock In \emph{International Conference on Learning Representations
  (ICLR)}, San Diego, CA, USA.

\bibitem[{Koehn et~al.(2007)Koehn, Hoang, Birch, Callison-Burch, Federico,
  Bertoldi, Cowan, Shen, Moran, Zens, Dyer, Bojar, Constantin, and
  Herbst}]{Koehn:2007:MOS:1557769.1557821}
Philipp Koehn, Hieu Hoang, Alexandra Birch, Chris Callison-Burch, Marcello
  Federico, Nicola Bertoldi, Brooke Cowan, Wade Shen, Christine Moran, Richard
  Zens, Chris Dyer, Ond\v{r}ej Bojar, Alexandra Constantin, and Evan Herbst.
  2007.
\newblock \href {http://dl.acm.org/citation.cfm?id=1557769.1557821} {Moses:
  Open source toolkit for statistical machine translation}.
\newblock In \emph{Proceedings of the 45th Annual Meeting of the ACL on
  Interactive Poster and Demonstration Sessions}, ACL '07, pages 177--180,
  Stroudsburg, PA, USA. Association for Computational Linguistics.

\bibitem[{Koncel-Kedziorski et~al.(2019)Koncel-Kedziorski, Bekal, Luan, Lapata,
  and Hajishirzi}]{rik_naacl19}
Rik Koncel-Kedziorski, Dhanush Bekal, Yi~Luan, Mirella Lapata, and Hannaneh
  Hajishirzi. 2019.
\newblock \href {https://doi.org/10.18653/v1/N19-1238} {{T}ext {G}eneration
  from {K}nowledge {G}raphs with {G}raph {T}ransformers}.
\newblock In \emph{Proceedings of the 2019 Conference of the North {A}merican
  Chapter of the Association for Computational Linguistics: Human Language
  Technologies, Volume 1 (Long and Short Papers)}, pages 2284--2293,
  Minneapolis, Minnesota. Association for Computational Linguistics.

\bibitem[{Konstas et~al.(2017)Konstas, Iyer, Yatskar, Choi, and
  Zettlemoyer}]{konsas_17}
Ioannis Konstas, Srinivasan Iyer, Mark Yatskar, Yejin Choi, and Luke
  Zettlemoyer. 2017.
\newblock \href {https://doi.org/10.18653/v1/P17-1014} {Neural amr:
  Sequence-to-sequence models for parsing and generation}.
\newblock In \emph{Proceedings of the 55th Annual Meeting of the Association
  for Computational Linguistics (Volume 1: Long Papers)}, pages 146--157,
  Vancouver, Canada. Association for Computational Linguistics.

\bibitem[{Li et~al.(2016)Li, Zemel, Brockschmidt, and Tarlow}]{Li2016GatedGS}
Yujia Li, Richard Zemel, Marc Brockschmidt, and Daniel Tarlow. 2016.
\newblock \href
  {https://www.microsoft.com/en-us/research/publication/gated-graph-sequence-neural-networks/}
  {Gated graph sequence neural networks}.
\newblock In \emph{Proceedings of the International Conference on Learning
  Representations (ICLR)}, San Juan, Puerto Rico.

\bibitem[{Liao et~al.(2018)Liao, Lebanoff, and Liu}]{liao-etal-2018-abstract}
Kexin Liao, Logan Lebanoff, and Fei Liu. 2018.
\newblock \href {https://www.aclweb.org/anthology/C18-1101} {Abstract meaning
  representation for multi-document summarization}.
\newblock In \emph{Proceedings of the 27th International Conference on
  Computational Linguistics}, pages 1178--1190, Santa Fe, New Mexico, USA.
  Association for Computational Linguistics.

\bibitem[{Marcheggiani and Perez~Beltrachini(2018)}]{marcheggiani-icnl18}
Diego Marcheggiani and Laura Perez~Beltrachini. 2018.
\newblock \href {https://www.aclweb.org/anthology/W18-6501} {Deep graph
  convolutional encoders for structured data to text generation}.
\newblock In \emph{Proceedings of the 11th International Conference on Natural
  Language Generation}, pages 1--9, Tilburg University, The Netherlands.
  Association for Computational Linguistics.

\bibitem[{Mille et~al.(2018)Mille, Belz, Bohnet, Graham, Pitler, and
  Wanner}]{mille-etal-2018-first}
Simon Mille, Anja Belz, Bernd Bohnet, Yvette Graham, Emily Pitler, and Leo
  Wanner. 2018.
\newblock \href {https://doi.org/10.18653/v1/W18-3601} {The first multilingual
  surface realisation shared task ({SR}{'}18): Overview and evaluation
  results}.
\newblock In \emph{Proceedings of the First Workshop on Multilingual Surface
  Realisation}, pages 1--12, Melbourne, Australia. Association for
  Computational Linguistics.

\bibitem[{Napoles et~al.(2012)Napoles, Gormley, and
  Van~Durme}]{Napoles:2012:AG:2391200.2391218}
Courtney Napoles, Matthew Gormley, and Benjamin Van~Durme. 2012.
\newblock \href {http://dl.acm.org/citation.cfm?id=2391200.2391218} {Annotated
  gigaword}.
\newblock In \emph{Proceedings of the Joint Workshop on Automatic Knowledge
  Base Construction and Web-scale Knowledge Extraction}, AKBC-WEKEX '12, pages
  95--100, Stroudsburg, PA, USA. Association for Computational Linguistics.

\bibitem[{Narayan and Gardent(2012)}]{narayan-gardent-2012-structure}
Shashi Narayan and Claire Gardent. 2012.
\newblock \href {https://www.aclweb.org/anthology/C12-1124} {Structure-driven
  lexicalist generation}.
\newblock In \emph{Proceedings of {COLING} 2012}, pages 2027--2042, Mumbai,
  India. The COLING 2012 Organizing Committee.

\bibitem[{Papineni et~al.(2002)Papineni, Roukos, Ward, and
  Zhu}]{Papineni:2002:BMA:1073083.1073135}
Kishore Papineni, Salim Roukos, Todd Ward, and Wei-Jing Zhu. 2002.
\newblock \href {https://doi.org/10.3115/1073083.1073135} {Bleu: A method for
  automatic evaluation of machine translation}.
\newblock In \emph{Proceedings of the 40th Annual Meeting on Association for
  Computational Linguistics}, ACL '02, pages 311--318, Stroudsburg, PA, USA.
  Association for Computational Linguistics.

\bibitem[{Pennington et~al.(2014)Pennington, Socher, and
  Manning}]{pennington-etal-2014-glove}
Jeffrey Pennington, Richard Socher, and Christopher Manning. 2014.
\newblock \href {https://doi.org/10.3115/v1/D14-1162} {{G}love: Global vectors
  for word representation}.
\newblock In \emph{Proceedings of the 2014 Conference on Empirical Methods in
  Natural Language Processing ({EMNLP})}, pages 1532--1543, Doha, Qatar.
  Association for Computational Linguistics.

\bibitem[{Pourdamghani et~al.(2016)Pourdamghani, Knight, and
  Hermjakob}]{pourdamghani-etal-2016-generating}
Nima Pourdamghani, Kevin Knight, and Ulf Hermjakob. 2016.
\newblock \href {https://doi.org/10.18653/v1/W16-6603} {Generating {E}nglish
  from abstract meaning representations}.
\newblock In \emph{Proceedings of the 9th International Natural Language
  Generation conference}, pages 21--25, Edinburgh, UK. Association for
  Computational Linguistics.

\bibitem[{Schlichtkrull et~al.(2018)Schlichtkrull, Kipf, Bloem, van~den Berg,
  Titov, and Welling}]{Schlichtkrull2018ModelingRD}
Michael~Sejr Schlichtkrull, Thomas~N. Kipf, Peter Bloem, Rianne van~den Berg,
  Ivan Titov, and Max Welling. 2018.
\newblock \href {https://doi.org/10.1007/978-3-319-93417-4\_38} {Modeling
  relational data with graph convolutional networks}.
\newblock In \emph{The Semantic Web - 15th International Conference, {ESWC}
  2018, Heraklion, Crete, Greece, June 3-7, 2018, Proceedings}, pages 593--607.

\bibitem[{Schuster and Paliwal(1997)}]{Schuster97bidirectionalrecurrent}
Mike Schuster and Kuldip~K. Paliwal. 1997.
\newblock \href {https://doi.org/10.1109/78.650093} {Bidirectional recurrent
  neural networks}.
\newblock \emph{IEEE Transactions on Signal Processing}, 45(11):2673--2681.

\bibitem[{See et~al.(2017)See, Liu, and Manning}]{see-etal-2017-get}
Abigail See, Peter~J. Liu, and Christopher~D. Manning. 2017.
\newblock \href {https://doi.org/10.18653/v1/P17-1099} {Get to the point:
  Summarization with pointer-generator networks}.
\newblock In \emph{Proceedings of the 55th Annual Meeting of the Association
  for Computational Linguistics (Volume 1: Long Papers)}, pages 1073--1083,
  Vancouver, Canada. Association for Computational Linguistics.

\bibitem[{Shieber et~al.(1990)Shieber, van Noord, Pereira, and
  Moore}]{shieber1990semantic}
Stuart~M. Shieber, Gertjan van Noord, Fernando C.~N. Pereira, and Robert~C.
  Moore. 1990.
\newblock \href {https://www.aclweb.org/anthology/J90-1004}
  {Semantic-head-driven generation}.
\newblock \emph{Computational Linguistics}, 16(1):30--42.

\bibitem[{Song et~al.(2019)Song, Gildea, Zhang, Wang, and
  Su}]{song-etal-2019-semantic}
Linfeng Song, Daniel Gildea, Yue Zhang, Zhiguo Wang, and Jinsong Su. 2019.
\newblock \href {https://doi.org/10.1162/tacl_a_00252} {Semantic neural machine
  translation using {AMR}}.
\newblock \emph{Transactions of the Association for Computational Linguistics},
  7:19--31.

\bibitem[{Song et~al.(2018)Song, Zhang, Wang, and Gildea}]{song-etal-acl2018}
Linfeng Song, Yue Zhang, Zhiguo Wang, and Daniel Gildea. 2018.
\newblock \href {https://www.aclweb.org/anthology/P18-1150} {A
  graph-to-sequence model for {AMR}-to-text generation}.
\newblock In \emph{Proceedings of the 56th Annual Meeting of the Association
  for Computational Linguistics (Volume 1: Long Papers)}, pages 1616--1626,
  Melbourne, Australia. Association for Computational Linguistics.

\bibitem[{Srivastava et~al.(2014)Srivastava, Hinton, Krizhevsky, Sutskever, and
  Salakhutdinov}]{Srivastava:2014:DSW:2627435.2670313}
Nitish Srivastava, Geoffrey Hinton, Alex Krizhevsky, Ilya Sutskever, and Ruslan
  Salakhutdinov. 2014.
\newblock \href {http://dl.acm.org/citation.cfm?id=2627435.2670313} {Dropout: A
  simple way to prevent neural networks from overfitting}.
\newblock \emph{Journal of Machine Learning Research}, 15(1):1929--1958.

\bibitem[{Veli{\v{c}}kovi{\'{c}} et~al.(2018)Veli{\v{c}}kovi{\'{c}}, Cucurull,
  Casanova, Romero, Li{\`{o}}, and Bengio}]{velickovic2018graph}
Petar Veli{\v{c}}kovi{\'{c}}, Guillem Cucurull, Arantxa Casanova, Adriana
  Romero, Pietro Li{\`{o}}, and Yoshua Bengio. 2018.
\newblock \href {https://openreview.net/forum?id=rJXMpikCZ} {{Graph Attention
  Networks}}.
\newblock In \emph{International Conference on Learning Representations},
  Vancouver, Canada.

\bibitem[{Weisfeiler and Lehman(1968)}]{weisfeiler}
Boris Weisfeiler and A.A. Lehman. 1968.
\newblock A reduction of a graph to a canonical form and an algebra arising
  during this reduction.
\newblock \emph{Nauchno-Technicheskaya Informatsia}, pages 12--16.

\bibitem[{Williams et~al.(2018)Williams, Nangia, and
  Bowman}]{williams-etal-2018-broad}
Adina Williams, Nikita Nangia, and Samuel Bowman. 2018.
\newblock \href {https://doi.org/10.18653/v1/N18-1101} {A broad-coverage
  challenge corpus for sentence understanding through inference}.
\newblock In \emph{Proceedings of the 2018 Conference of the North {A}merican
  Chapter of the Association for Computational Linguistics: Human Language
  Technologies, Volume 1 (Long Papers)}, pages 1112--1122, New Orleans,
  Louisiana. Association for Computational Linguistics.

\bibitem[{Xu et~al.(2019)Xu, Hu, Leskovec, and Jegelka}]{xu2018how}
Keyulu Xu, Weihua Hu, Jure Leskovec, and Stefanie Jegelka. 2019.
\newblock \href {https://openreview.net/forum?id=ryGs6iA5Km} {How powerful are
  graph neural networks?}
\newblock In \emph{International Conference on Learning Representations}, New
  Orleans, LA, USA.

\end{thebibliography}
\bibliographystyle{acl_natbib}

\clearpage

\appendix

\section{Generated Sentences}

Table~\ref{tab:example1} shows three examples with sentences generated by {\fontfamily{qcr}\selectfont S2S}, Song et al. (2018), {\fontfamily{qcr}\selectfont G2S-GGNN}, and the reference sentence ({\fontfamily{qcr}\selectfont GOLD}).

\begin{table*}[b]
\renewcommand{\arraystretch}{1.2}
\linespread{1}\selectfont\centering
 \begin{tabular}{p{2.75cm} p{12.5cm}} 
 \hline
{\fontfamily{qcr}\selectfont GOLD} & {\small I don't want to be miserable anymore and the longer he is around the more miserable I will be.
}   \\
{\fontfamily{qcr}\selectfont S2S} & {\small If he was in longer longer, I don't want to miserable and more miserable.}\\
Song et. al (2018) & {\small I don't want to be miserable anymore, and when he is around longer, I'm a miserable miserable.} \\
{\fontfamily{qcr}\selectfont G2S-GGNN} & {\small I don't want to be miserable anymore, and would be more miserable if he was around longer.} \\
 \hline
 & \\[0.2ex]
  \hline
{\fontfamily{qcr}\selectfont GOLD} & {\small Colombia is the source of much of the cocaine and heroin sold in the United States.}   \\
{\fontfamily{qcr}\selectfont S2S} & {\small  Colombia is a source of cocaine, much of cocaine and heroin sales in the United States.}\\
Song et. al (2018) & {\small Colombia is a source of much of much of cocaine and heroin in the United States.} \\
{\fontfamily{qcr}\selectfont G2S-GGNN} & {\small Colombia is a source of much cocaine and heroin and heroin sold in the United States.} \\
 \hline
 & \\
  \hline
{\fontfamily{qcr}\selectfont GOLD} & {\small Discussions between Lula da Silva and Thabo Mbeki would also address new threats to international security such as terrorism, drugs, illegal weapons trafficking and aids.}   \\
{\fontfamily{qcr}\selectfont S2S} & {\small  Thabo da Silva has also addressed Thabo Mbeki to discuss new threats such as terrorism, drugs, illegal weapons trafficking and aids in international security.}\\
Song et. al (2018) & {\small Lula da Silva's discussion with Thabo also addressed a new threat against Thabo Mbeki and aids, drugs, illegal weapons and illegal weapons of weapon.} \\
{\fontfamily{qcr}\selectfont G2S-GGNN} & {\small Lula da Silva's discussion with Thabo da Silva also addressed new threat such as terrorism, drugs, illegal weapons trafficking and aids.} \\
 \hline
\end{tabular}
\caption{Examples of generated sentences. GOLD refers to the reference sentence.}
\label{tab:example1}
\end{table*}

\section{Human Evaluation Setup}

\begin{itemize}
    \item For each quality evaluation task (meaning similarity and readability), we independently sampled 100 generated sentences for each model.
    \item We created separate HITs for meaning similarity and readability evaluations. Each HIT contains 10 sentences.
    \item We paid \$0.15 per HIT, employing five workers on each. For qualification, workers were required to have over 1000 approved HITs.
    \item We applied a quality control step. We removed workers who do not achieve a minimum threshold in sentences with known scores. 
    
\end{itemize}

\end{document}